\documentclass[final]{article}
\usepackage{corl2020}

\usepackage{amsmath}
\usepackage{amssymb}
\usepackage{graphicx}
\usepackage{tabularx}
\usepackage{subcaption}
\usepackage{textcomp}
\usepackage{tikz}
\usetikzlibrary{calc}
\usetikzlibrary{arrows}
\usetikzlibrary{shapes}
\usepackage{pgfplots}
\pgfplotsset{compat=newest}
\usetikzlibrary{plotmarks}
\usetikzlibrary{arrows.meta}
\usepgfplotslibrary{patchplots}
\usepackage{grffile}

\def\onedot{.\ }
 
\def\eg{\emph{e.g}\onedot} 
\def\etal{\emph{et al}\onedot}
 
\def\ie{\emph{i.e}\onedot}

\DeclareMathOperator*{\argmin}{argmin}
\newcommand{\PLH}{{\mkern-2mu\times\mkern-2mu}}

\graphicspath{{figures/}}

\title{Unsupervised Metric Relocalization\\
Using Transform Consistency Loss\vspace{-0.25em}}

\author{
  Mike~Kasper\\
  University of Colorado, USA\\
  \texttt{michael.kasper@colorado.edu} \\
  \And
  Fernando~Nobre\\
  Amazon, USA\\
  \texttt{fnobre@amazon.com} \\
  \And
  Christoffer~Heckman\\
  University of Colorado, USA\\
  \texttt{christoffer.heckman@colorado.edu} \\
  \And
  Nima~Keivan\\
  Amazon, USA\\
  \texttt{keivan@amazon.com} \\
}

\begin{document}
\maketitle
\vspace{-1em}


\begin{abstract}
  Training networks to perform metric relocalization traditionally requires
  accurate image correspondences. In practice, these are obtained by restricting
  domain coverage, employing additional sensors, or capturing large multi-view
  datasets. We instead propose a self-supervised solution, which exploits a key
  insight: localizing a query image within a map should yield the same absolute
  pose, regardless of the reference image used for registration. Guided by this
  intuition, we derive a novel \textit{transform consistency loss}. Using this
  loss function, we train a deep neural network to infer dense feature and
  saliency maps to perform robust metric relocalization in dynamic environments.
  We evaluate our framework on synthetic and real-world data, showing our
  approach outperforms other supervised methods when a limited amount of
  ground-truth information is available.
\end{abstract}



\section{Introduction}
\label{sec:loc-introduction}

Visual relocalization refers to the task of registering a query image to an
existing map, effectively seeking an answer to the question, ``have I been here
before?'' This is a critical problem in autonomous robot navigation
\cite{MurArtal-TR-2017, Qin-TR-2018, Gao-IROS-2018}. Relocalization may be
limited to image retrieval, where the query image is assumed to share the same
pose as the matched reference image \cite{Noh-ICCV-2017, Garg-ARXIV-2018,
Arandjelovic-TPAMI-2018}. More challenging still is the task of metrically
refining this pose by estimating a 6 degree-of-freedom (DoF) offset between the
two images \cite{Gomez-ICRA-Ojeda2018, VonStumberg-RAL-2019}. In this work, we
focus only on the latter problem.

As the reference and query images may be separated by an arbitrary amount of
time, large changes in visual appearance can inhibit accurate relocalization.
Hand-engineered feature descriptors have proven insufficient in extremely
dynamic environments \cite{Verdie-CVPR-2015, Yi-ECCV-2016,
VonStumberg-RAL-2019}. Consequently, recent approaches have pursued the use of
neural networks to learn more robust image representations \cite{Choy-NIPS-2016,
Schmidt-ICRA-2017, VonStumberg-RAL-2019}. However, these network architectures
are typically trained with ground-truth image correspondences, requiring the
scene geometry and camera poses to be known.


\begin{figure}
  \centering
  \begin{subfigure}{0.24\textwidth}
    \includegraphics[width=\textwidth,height=0.60\textwidth,
    trim=20 0 20 0, clip]{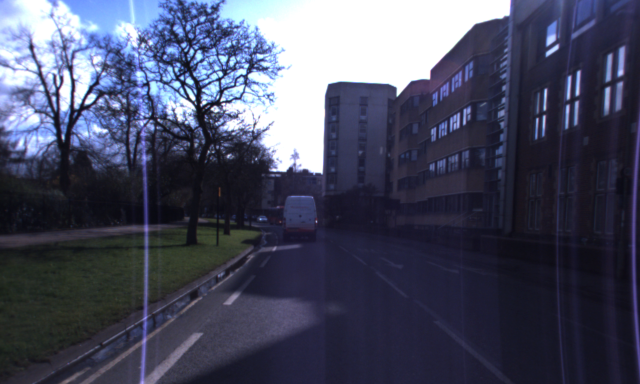}
  \end{subfigure}
  \begin{subfigure}{0.24\textwidth}
    \includegraphics[width=\textwidth,height=0.60\textwidth,
    trim=20 0 20 0, clip]{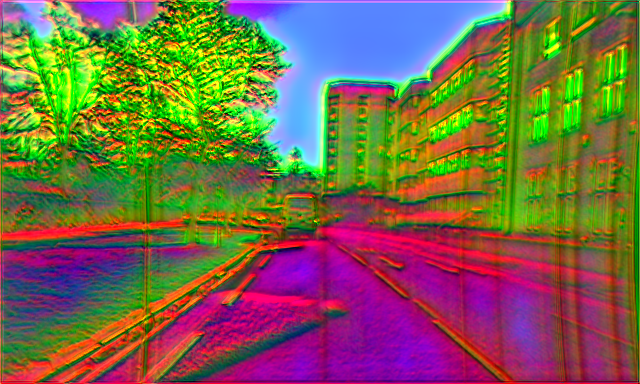}
  \end{subfigure}
  \begin{subfigure}{0.24\textwidth}
    \includegraphics[width=\textwidth,height=0.60\textwidth,
    trim=20 0 20 0, clip]{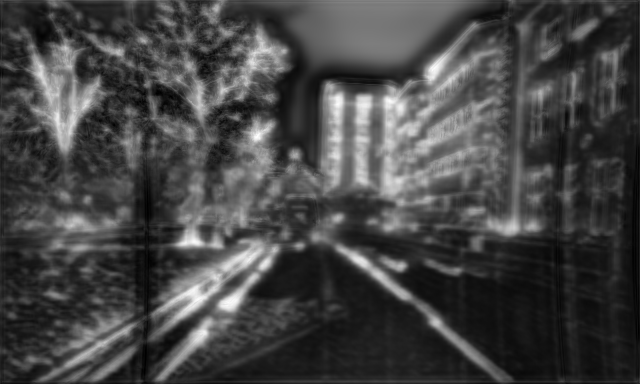}
  \end{subfigure}
  \begin{subfigure}{0.24\textwidth}
    \includegraphics[width=\textwidth,height=0.60\textwidth,
    trim=20 0 20 0, clip]{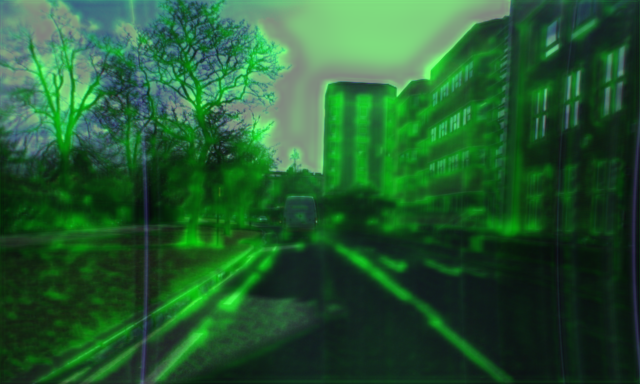}
  \end{subfigure}\\[0.25em]
  \begin{subfigure}{0.24\textwidth}
    \includegraphics[width=\textwidth,height=0.60\textwidth,
    trim=20 0 20 0, clip]{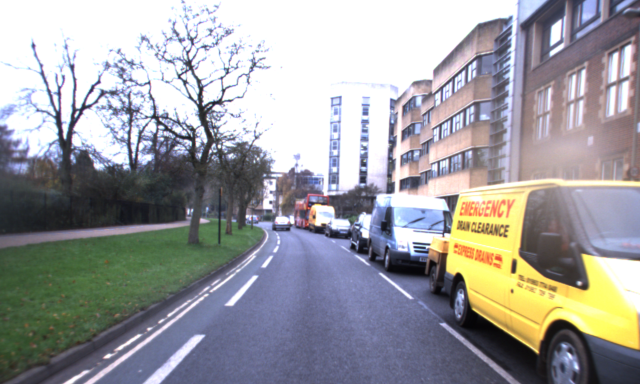}
    \vspace{-1.65em} \caption*{\bf\scriptsize Input Image}
  \end{subfigure}
  \begin{subfigure}{0.24\textwidth}
    \includegraphics[width=\textwidth,height=0.60\textwidth,
    trim=20 0 20 0, clip]{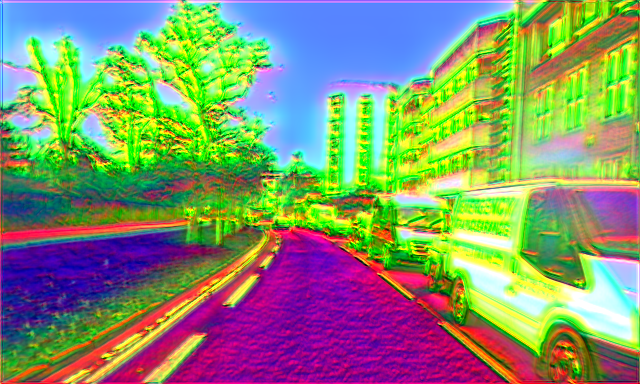}
    \vspace{-1.65em} \caption*{\bf\scriptsize Feature Map}
  \end{subfigure}
  \begin{subfigure}{0.24\textwidth}
    \includegraphics[width=\textwidth,height=0.60\textwidth,
    trim=20 0 20 0, clip]{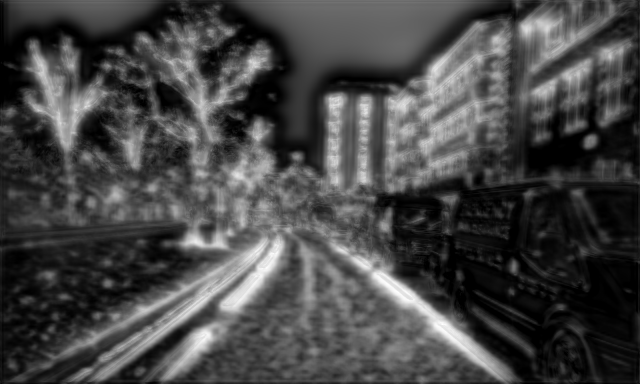}
    \vspace{-1.65em} \caption*{\bf\scriptsize Saliency Map}
  \end{subfigure}
  \begin{subfigure}{0.24\textwidth}
    \includegraphics[width=\textwidth,height=0.60\textwidth,
    trim=20 0 20 0, clip]{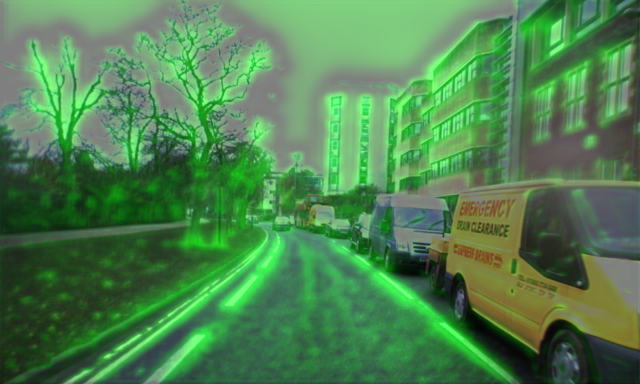}
    \vspace{-1.65em} \caption*{\bf\scriptsize Saliency Overlay}
  \end{subfigure}
  \caption[Example feature and saliency maps]{Images of the same location
    taken at different times and their resulting feature and saliency maps. For
    visualization purposes, we fuse all pyramid levels together using PCA for
    feature maps and a weighted average for saliency maps. Note the similarity
    of the feature maps, despite large discrepancies in the input images, and
    how the saliency maps successfully mask out dynamic objects.}
  \label{fig:teaser}
\end{figure}


Acquiring these ground-truth data in the real world is non-trivial. Simpler
solutions sacrifice domain coverage by restricting camera placement or
synthetically rendering alternate views. Better coverage of the target domain is
achieved via additional sensing or large image sets of the same scene. These
requirements make one-time data capture burdensome and online capture with
lightweight robot platforms nearly impossible, thus limiting our ability to
adapt to novel environments.

To circumvent these problems, we instead propose a solution that does not
require ground-truth image correspondences while still achieving full coverage
of the target domain. This permits an unmodified version of our robot platform
to collect training data while deployed. With our framework, training samples
can be captured periodically, when deemed both prudent and convenient by the
robot. To make correspondence-free training possible, we propose a novel loss
function inspired by a key insight: when localizing a query image within a map,
the same absolute pose should be obtained, regardless of the reference image
used for registration. That is, given two reference images in the same world
frame, aligning a query image to either one should yield the same global pose.

Guided by this intuition, we develop a novel \textit{transform consistency
loss}, which operates over three images: two reference images, whose relative
pose is known, and a single query image, whose relative pose is unknown.
Employing this loss function, we train a network to infer dense feature and
saliency maps, as seen in Figure~\ref{fig:teaser}. These can then be used to
perform direct image registration that is robust to dynamic objects and
illumination. The novel contributions of this work are:
\begin{itemize}
  \item A transform consistency loss function for unsupervised learning.
  \item A network architecture for joint feature and saliency map inference.
\end{itemize}
By removing many of the requirements on data acquisition, we can learn from a
much wider range of real-world images. Consequently, our framework significantly
outperforms other state-of-the-art, supervised methods when trained on datasets
of limited size and scope.


\section{Related Work}
\label{sec:loc-related-work}

Traditionally, hand-engineered features have been used for metric relocalization
\cite{Lowe-IJCV-2004, Bay-ECCV-2006, Rublee-ICCV-2011, Dalal-NONE-2005}. They
are designed to be invariant to changes in scale, orientation, and illumination.
ORB-SLAM \cite{MurArtal-TR-2017} is an example framework that employs such
features. It first retrieves image candidates via a place recognition system
based on bags of binary words \cite{Galvez-Lopez-TR-2012}, and then estimates a
6-DoF offset by minimizing the reprojection error between the image coordinates
matched using ORB descriptors \cite{Rublee-ICCV-2011}. While hand-engineered
features work well for visual odometry, they have proven inadequate for
relocalization in extremely dynamic environments
\cite{Yi-ECCV-2016,VonStumberg-RAL-2019}.

To address the deficiencies of hand-engineered features, approaches relying on
alternative image representations generated by neural networks have risen in
popularity. These frameworks have been used to perform sparse keypoint detection
and description \cite{Verdie-CVPR-2015, Yi-ECCV-2016, Detone-CVPR-2018,
Revaud-NIPS-2019} as well as dense image registration \cite{Tang-ARXIV-2018,
Han-ARXIV-2018, Lv-CVPR-2018, VonStumberg-RAL-2019}. However, current solutions
assume ground-truth correspondences are available at training time in order to
compare multiple observations of the same point. Example loss functions include
\emph{contrastive loss} \cite{Hadsell-CVPR-2006} and \emph{cosine similarity}
\cite{Revaud-NIPS-2019}.

Several methods for obtaining these ground-truth correspondences have been
proposed. Perhaps the simplest approach of all is to use simulated data, as we
have perfect knowledge of scene geometry and camera poses
\cite{Detone-CVPR-2018, VonStumberg-RAL-2019}. However, the viability of
\textit{sim2real} transfer learning greatly depends on the target
environment. In general, we cannot enumerate an exhaustive list of the expected
visual phenomena, let alone render them with a sufficient level of fidelity.
Consequently, most approaches employ real-world data and therefore must directly
tackle the correspondence problem.

One solution is to apply known transformations (\eg color shifts, homographic
warps) to real-world images \cite{Detone-CVPR-2018, Revaud-NIPS-2019}. However,
this still relies on synthetic data and therefore runs the risk of
insufficiently sampling the target domain. Alternatively, Verdie \etal propose
using a stationary camera to observe the same scene under varying lighting and
weather conditions \cite{Verdie-CVPR-2015}. This makes the correspondence
problem trivial, as the same pixel corresponds to the same world point in every
image, but fails to capture visual artifacts produced by camera motion. In a
similar vein, Liang \etal employ RTK positioning to obtain images captured from
the same viewpoint \cite{Liang-TASE-2019}. Not only does this require additional
hardware, it also constrains the relative camera poses to be identity.
In \cite{VonStumberg-RAL-2019}, the authors employ a visual-odometry framework
to automate keypoint selection and depth estimation, but require lidar data
or a motion-capture system to obtain the inter-sequence transforms.

A popular alternative to sensor-based solutions is Structure-from-Motion (SfM)
\cite{Yi-ECCV-2016, VonStumberg-RAL-2019, Dusmanu-CVPR-2019, Revaud-NIPS-2019}.
This involves processing large image sets of the \textit{same} scene to build a
cohesive model of the surface geometry and camera poses. This is arguably the
most general solution to the correspondence problem, as it even works on random
image collections scraped from internet. However, SfM cannot be used to extract
correspondences from sparsely sampled scenes. It also assumes that the images
are similar enough that existing methods can perform registration.

More recently, Schmidt \etal propose a self-supervised approach for learning
robust feature descriptors \cite{Schmidt-ICRA-2017}. Training data is sourced
from multiple, independent 3D reconstructions of a single subject, built using a
3D reconstruction pipeline (\ie DynamicFusion \cite{Newcombe-CVPR-2015}). They
then train solely on \emph{intra}-sequence correspondences. Despite not
training on \emph{inter}-sequence correspondences, the authors empirically find
the inferred features map well across each reconstruction. However, their
trained model only serves for a single subject. It also requires large amounts
of data to be captured and processed, which is problematic if we expect this to
be done \textit{in situ} by a deployed mobile agent.

To circumvent the aforementioned challenges for obtaining ground-truth image
correspondences, we propose dropping the requirement altogether. In the next
section, we present our novel loss function, which assesses the consistency of
estimated 6-DoF camera poses. It leverages data already available in most visual
localization frameworks, making it suitable for a much broader range of
applications.


\section{Method}
\label{sec:method}

\begin{figure}
  \centering
  \begin{subfigure}{0.40\textwidth}
    \centering
    \scalebox{0.65}{\begin{tikzpicture}[fill=blue!20]
      \coordinate (Tr0) at (1.0, 1.0);
      \coordinate (Tr1) at (4.8, 3.6);
      \coordinate (Tt)  at (3.6, 0.3);
      \coordinate (lm0) at (9.0, 3.3);
      \draw[thick,dotted,black] (Tr1)--(lm0);
      \draw[thick,dotted,black] (Tr1)--(lm0);
      \draw[thick,dotted,black] (Tr0)--(lm0);
      \draw[thick,dotted,black] (Tt)--(lm0);
      \draw[thick,solid,black]  (Tr0)..controls +( 0.1,  2.5) and +(-1.1,  0.3)..(Tr1);
      \draw[thick,solid,black] (Tr0)..controls +( 0.5, -0.7) and +(-1.0, -0.3)..(Tt);
      \draw[thick,solid,black] (Tr1)..controls +(-1.3, -1.0) and +( 0.0,  1.0)..(Tt);
      \path (Tr0) node(a) [isosceles triangle,rotate=210,draw,fill,minimum height=5em, anchor=east]{}
            (Tr1) node(b) [isosceles triangle,rotate=180,draw,fill,minimum height=5em, anchor=east]{}
            (Tt)  node(c) [isosceles triangle,rotate=215,draw,fill=red!20,minimum height=5em, anchor=east]{}
            (lm0) node(d) [star,draw,fill=yellow!20]{};
      \node at (0.7,  0.8) {\Large $T_{r_0}$};
      \node at (4.8,  4.0) {\Large $T_{r_1}$};
      \node at (3.6, -0.1) {\Large $T_q$};
      \node at (9.05,  3.9) {\Large $p$};
    \end{tikzpicture}}
  \end{subfigure}
  \begin{subfigure}{0.05\textwidth}
    \hfill
  \end{subfigure}
  \begin{subfigure}{0.40\textwidth}
    \centering
    \scalebox{0.65}{\begin{tikzpicture}[fill=blue!20]
      \coordinate (Tr0) at (1.0, 1.0);
      \coordinate (Tr1) at (4.8, 3.6);
      \coordinate (Tt)  at (3.0, 0.3);
      \coordinate (Ts)  at (5.4, 0.8);
      \coordinate (lm0) at (9.0, 3.3);
      \draw[thick,dotted,black] (Tr1)--(lm0);
      \draw[thick,dotted,black] (Tr1)--(lm0);
      \draw[thick,dotted,black] (Tr0)--(lm0);
      \draw[thick,dotted,black] (Tt)--(lm0);
      \draw[thick,dotted,black] (Ts)--(lm0);
      \draw[thick,solid,black]  (Tr0)..controls +( 0.1,  2.5) and +(-1.1,  0.3)..(Tr1);
      \draw[thick,dashed,black] (Tr0)..controls +( 0.5, -0.7) and +(-1.0, -0.1)..(Tt);
      \draw[thick,dashed,black] (Tr1)..controls +(-0.6, -1.5) and +(-0.4,  0.5)..(Ts);
      \path (Tr0) node(a) [isosceles triangle,rotate=210,draw,fill,minimum height=5em, anchor=east]{}
            (Tr1) node(b) [isosceles triangle,rotate=180,draw,fill,minimum height=5em, anchor=east]{}
            (Tt)  node(c) [isosceles triangle,rotate=205,draw,fill=red!20,minimum height=5em, anchor=east]{}
            (Ts)  node(d) [isosceles triangle,rotate=220,draw,fill=red!20,minimum height=5em, anchor=east]{}
            (lm0) node(e) [star,draw,fill=yellow!20]{};
      \node at (0.7,  0.8) {\Large $T_{r_0}$};
      \node at (4.8,  4.0) {\Large $T_{r_1}$};
      \node at (3.0, -0.1) {\Large $T_{q_0}$};
      \node at (5.4,  0.4) {\Large $T_{q_1}$};
      \node at (9.05,  3.9) {\Large $p$};
    \end{tikzpicture}}
  \end{subfigure}
  \caption[Transform consistency graph]{A visualization of transform
    consistency. On the left, we see the ground-truth configuration of three
    frames $r_0, r_1, q$. On the right, we see the result of
    \emph{independently} aligning the query image $q$ with each reference image
    $r_0, r_1$. We seek a robust representation of the mutually visible points
    $p$ as to minimize the distance between the estimated transforms
    $T_{r_0,q_0}$ and $T_{r_0,q_1}$.}
  \label{fig:graph}
\end{figure}
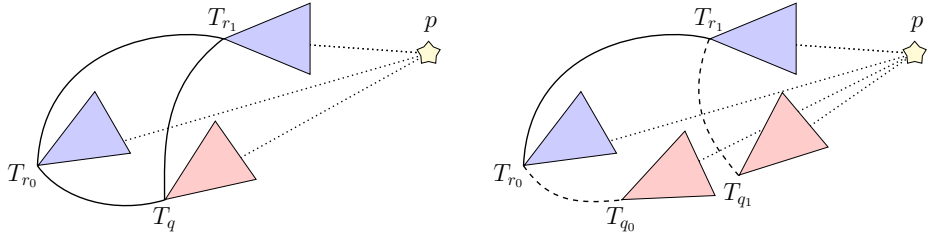

Our problem formulation is largely inspired by \emph{photometric consistency}.
In short, photometric consistency uses the inferred model of the scene to render
a synthetic image under different conditions (\eg camera pose). The fidelity of
the model can then assessed by comparing this rendering with an actual image
captured under those same conditions. Photometric consistency has been used to
learn depth from both monocular \cite{Zhou-CVPR-2017, Godard-CVPR-2017} and
stereo images \cite{Garg-ECCV-2016}.

Employing photometric consistency requires a static scene, where the brightness
constancy assumption holds (although subtle illumination changes may be handled
\cite{Khot-CVPR-2019}). Such assumptions, however, cannot be made for
relocalization. We therefore substitute photometric consistency for what we
refer to as \textit{transform consistency}. Here we use the inferred image
features to \textit{independently} register a query image with two reference
images, whose relative pose is already known. We then evaluate our network by
comparing the resulting transforms in a shared world frame. This concept is
illustrated in Figure~\ref{fig:graph}. To employ this method, we make the
following assumptions:
\begin{itemize}
  \item Dense depth maps are provided for both reference images.
  \item An accurate relative transform between the reference images is provided.
  \item The query image's visual-field sufficiently overlaps with both reference
    images.
\end{itemize}
In practice, we obtain depth maps using stereo-vision
\cite{Hirschmuller-CVPR-2005}, although an RGB-D sensor would also suffice. To
ensure the accuracy of the relative transforms, we sample both references frames
from a small spatio-temporal window of a captured video sequence. Off-the-shelf
visual odometry pipelines can therefore reliably attain translation errors
within 1\% of the distance traveled \cite{MurArtal-TR-2017, Qin-TR-2018}.
Finally, we uphold the \emph{overlapping visual-field} assumption via a GPS. We
can also leverage an image-retrieval system or crude registrations to a
topological map \cite{Churchill-ICRA-2012}. Most vision-based, autonomous mobile
platforms already satisfy these three requirements. Consequently, they can
capture training data in the field, without additional sensing or significant
computational overhead. In the sections that follow, we present our network
architecture and formally define our objective function.


\subsection{Network Architecture}
\label{sec:loc-network-architecture}

\begin{figure}
  \centering
  \scalebox{0.85}{\begin{tikzpicture}

    \draw[white] (-1.10, -1.58) rectangle (13.03, 3.90);

    \draw [-latex, line width=0.3mm] (-0.2, 3.35) -- (11.96, 3.35);

    \draw [-latex, line width=0.3mm] (0.8, 2.50) -- (10.70, 2.50);

    \draw [-latex, line width=0.3mm] (2.5, 1.88) -- (8.75, 1.88);

    \draw [-latex, line width=0.3mm] (4.5, 1.37) -- (6.80, 1.37);

    \tikzset{shift={(8.35, 0.00)}};
    \draw [red!80!black,-latex,line width=0.3mm] (0.00, 0.00) -- (0.00, -0.80);
    \draw (0, -1.00) node[align=center]{\footnotesize $F_3, S_3$};
    \tikzset{shift={(-8.35, 0.00)}};

    \tikzset{shift={(+10.29, 0.00)}};
    \draw [red!80!black,-latex,line width=0.3mm] (0.00, 0.00) -- (0.00, -0.80);
    \draw (0, -1.00) node[align=center]{\footnotesize $F_2, S_2$};
    \tikzset{shift={(-10.29, 0.00)}};

    \tikzset{shift={(+11.61, 0.00)}};
    \draw [red!80!black,-latex,line width=0.3mm] (0.00, 0.00) -- (0.00, -0.80);
    \draw (0, -1.00) node[align=center]{\footnotesize $F_1, S_1$};
    \tikzset{shift={(-11.61, 0.00)}};

    \tikzset{shift={(+12.63, 0.00)}};
    \draw [red!80!black,-latex,line width=0.3mm] (0.00, 0.00) -- (0.00, -0.80);
    \draw (0, -1.00) node[align=center]{\footnotesize $F_0, S_0$};
    \tikzset{shift={(-12.63, 0.00)}};

    \tikzset{shift={(-0.60, -0.30)}};
    \draw[black] (0.0, 0.0) rectangle (6.36, -1.28);

      \tikzset{shift={(+0.10, -0.10)}};
      \fill[black!15!white] (0.00, 0.00) rectangle (0.30, -0.30);
      \draw[black] (0.00, -0.00) rectangle (0.30, -0.30);
      \draw (1.67, -0.16) node[align=left]{\footnotesize VGG Convolution};
      \tikzset{shift={(-0.10, +0.10)}};

      \tikzset{shift={(+0.10, -0.50)}};
      \fill[red!30!white] (0.00, 0.00) rectangle (0.30, -0.30);
      \draw[black] (0.00, -0.00) rectangle (0.30, -0.30);
      \draw (1.44, -0.175) node[align=left]{\footnotesize $2\times2$ Maxpool};
      \tikzset{shift={(-0.10, +0.10)}};

      \tikzset{shift={(+0.10, -0.50)}};
      \draw [-latex,line width=0.3mm] (0.00, -0.15)--(0.325, -0.15);
      \draw (1.40, -0.13) node[align=left]{\footnotesize Concatenation};
      \tikzset{shift={(-0.10, +0.90)}};

      \tikzset{shift={(+3.35, -0.10)}};
      \fill[green!30!white] (0.00, 0.00) rectangle (0.30, -0.30);
      \draw[black] (0.00, -0.00) rectangle (0.30, -0.30);
      \draw (1.07, -0.17) node[align=left]{\footnotesize Upsample};
      \tikzset{shift={(-3.35, +0.10)}};

      \tikzset{shift={(+3.35, -0.50)}};
      \fill[blue!30!white] (0.00, 0.00) rectangle (0.30, -0.30);
      \draw[black] (0.00, -0.00) rectangle (0.30, -0.30);
      \draw (1.67, -0.15) node[align=left]{\footnotesize$3\times3$ Convolution};
      \tikzset{shift={(-3.35, +0.10)}};

      \tikzset{shift={(+3.35, -0.50)}};
      \draw [red!80!black,-latex,line width=0.3mm] (0.00, -0.15)--(0.325, -0.15);
      \draw (1.11, -0.13) node[align=left]{\footnotesize Prediction};
      \tikzset{shift={(-3.35, +0.80)}};

    \tikzset{shift={(+0.60, +0.30)}};


    \tikzset{shift={(-0.80, 0.00)}};
    \fill[white] (0.00, 0.00) rectangle (0.10, 4.00);
    \draw[black] (0.00, 0.00) rectangle (0.10, 4.00);

    \tikzset{shift={(0.15, 0.00)}};
    \fill[black!15!white] (0.00, 0.00) rectangle (0.20, 4.00);
    \draw[black] (0.00, 0.00) rectangle (0.20, 4.00);

    \tikzset{shift={(0.25, 0.00)}};
    \fill[black!15!white] (0.00, 0.00) rectangle (0.20, 4.00);
    \draw[black] (0.00, 0.00) rectangle (0.20, 4.00);


    \tikzset{shift={(0.40, 0.00)}};
    \fill[red!30!white] (0.00, 0.00) rectangle (0.20, 3.00);
    \draw[black] (0.00, 0.00) rectangle (0.20, 3.00);

    \tikzset{shift={(0.25, 0.00)}};
    \fill[black!15!white] (0.00, 0.00) rectangle (0.28, 3.00);
    \draw[black] (0.00, 0.00) rectangle (0.28, 3.00);

    \tikzset{shift={(0.33, 0.00)}};
    \fill[black!15!white] (0.00, 0.00) rectangle (0.28, 3.00);
    \draw[black] (0.00, 0.00) rectangle (0.28, 3.00);


    \tikzset{shift={(0.48, 0.00)}};
    \fill[red!30!white] (0.00, 0.00) rectangle (0.28, 2.25);
    \draw[black] (0.00, 0.00) rectangle (0.28, 2.25);

    \tikzset{shift={(0.33, 0.00)}};
    \fill[black!15!white] (0.00, 0.00) rectangle (0.40, 2.25);
    \draw[black] (0.00, 0.00) rectangle (0.40, 2.25);

    \tikzset{shift={(0.45, 0.00)}};
    \fill[black!15!white] (0.00, 0.00) rectangle (0.40, 2.25);
    \draw[black] (0.00, 0.00) rectangle (0.40, 2.25);

    \tikzset{shift={(0.45, 0.00)}};
    \fill[black!15!white] (0.00, 0.00) rectangle (0.40, 2.25);
    \draw[black] (0.00, 0.00) rectangle (0.40, 2.25);


    \tikzset{shift={(0.60, 0.00)}};
    \fill[red!30!white] (0.00, 0.00) rectangle (0.40, 1.69);
    \draw[black] (0.00, 0.00) rectangle (0.40, 1.69);

    \tikzset{shift={(0.45, 0.00)}};
    \fill[black!15!white] (0.00, 0.00) rectangle (0.40, 1.69);
    \draw[black] (0.00, 0.00) rectangle (0.40, 1.69);

    \tikzset{shift={(0.45, 0.00)}};
    \fill[black!15!white] (0.00, 0.00) rectangle (0.40, 1.69);
    \draw[black] (0.00, 0.00) rectangle (0.40, 1.69);

    \tikzset{shift={(0.45, 0.00)}};
    \fill[black!15!white] (0.00, 0.00) rectangle (0.40, 1.69);
    \draw[black] (0.00, 0.00) rectangle (0.40, 1.69);


    \tikzset{shift={(0.60, 0.00)}};
    \fill[red!30!white] (0.00, 0.00) rectangle (0.40, 1.27);
    \draw[black] (0.00, 0.00) rectangle (0.40, 1.27);

    \tikzset{shift={(0.45, 0.00)}};
    \fill[black!15!white] (0.00, 0.00) rectangle (0.40, 1.27);
    \draw[black] (0.00, 0.00) rectangle (0.40, 1.27);

    \tikzset{shift={(0.45, 0.00)}};
    \fill[black!15!white] (0.00, 0.00) rectangle (0.40, 1.27);
    \draw[black] (0.00, 0.00) rectangle (0.40, 1.27);

    \tikzset{shift={(0.45, 0.00)}};
    \fill[black!15!white] (0.00, 0.00) rectangle (0.40, 1.27);
    \draw[black] (0.00, 0.00) rectangle (0.40, 1.27);


    \tikzset{shift={(0.60, 0.00)}};
    \fill[green!30!white] (0.00, 0.00) rectangle (0.40, 1.69);
    \draw[black] (0.00, 0.00) rectangle (0.40, 1.69);

    \tikzset{shift={(0.45, 0.00)}};
    \fill[blue!30!white] (0.00, 0.00) rectangle (0.40, 1.69);
    \draw[black] (0.00, 0.00) rectangle (0.40, 1.69);

    \tikzset{shift={(0.45, 0.00)}};
    \fill[blue!30!white] (0.00, 0.00) rectangle (0.40, 1.69);
    \draw[black] (0.00, 0.00) rectangle (0.40, 1.69);

    \tikzset{shift={(0.45, 0.00)}};
    \fill[blue!30!white] (0.00, 0.00) rectangle (0.40, 1.69);
    \draw[black] (0.00, 0.00) rectangle (0.40, 1.69);


    \tikzset{shift={(0.60, 0.00)}};
    \fill[green!30!white] (0.00, 0.00) rectangle (0.40, 2.25);
    \draw[black] (0.00, 0.00) rectangle (0.40, 2.25);

    \tikzset{shift={(0.45, 0.00)}};
    \fill[blue!30!white] (0.00, 0.00) rectangle (0.40, 2.25);
    \draw[black] (0.00, 0.00) rectangle (0.40, 2.25);

    \tikzset{shift={(0.45, 0.00)}};
    \fill[blue!30!white] (0.00, 0.00) rectangle (0.40, 2.25);
    \draw[black] (0.00, 0.00) rectangle (0.40, 2.25);

    \tikzset{shift={(0.45, 0.00)}};
    \fill[blue!30!white] (0.00, 0.00) rectangle (0.40, 2.25);
    \draw[black] (0.00, 0.00) rectangle (0.40, 2.25);


    \tikzset{shift={(0.60, 0.00)}};
    \fill[green!30!white] (0.00, 0.00) rectangle (0.40, 3.00);
    \draw[black] (0.00, 0.00) rectangle (0.40, 3.00);

    \tikzset{shift={(0.45, 0.00)}};
    \fill[blue!30!white] (0.00, 0.00) rectangle (0.28, 3.00);
    \draw[black] (0.00, 0.00) rectangle (0.28, 3.00);

    \tikzset{shift={(0.33, 0.00)}};
    \fill[blue!30!white] (0.00, 0.00) rectangle (0.28, 3.00);
    \draw[black] (0.00, 0.00) rectangle (0.28, 3.00);


    \tikzset{shift={(0.48, 0.00)}};
    \fill[green!30!white] (0.00, 0.00) rectangle (0.28, 4.00);
    \draw[black] (0.00, 0.00) rectangle (0.28, 4.00);

    \tikzset{shift={(0.33, 0.00)}};
    \fill[blue!30!white] (0.00, 0.00) rectangle (0.20, 4.00);
    \draw[black] (0.00, 0.00) rectangle (0.20, 4.00);

    \tikzset{shift={(0.25, 0.00)}};
    \fill[blue!30!white] (0.00, 0.00) rectangle (0.20, 4.00);
    \draw[black] (0.00, 0.00) rectangle (0.20, 4.00);

  \end{tikzpicture}}
  \caption[Network architecture]{An overview of our network architecture. The
    contractive part of the network leverages the pretrained convolutional
    layers of VGG-16 \cite{Simonyan-ARXIV-2015}, which remain constant. We
    independently pass each input image through the network to obtain a
    16-channel feature map $F$ and a single-channel saliency map $S$ at each
    level of the image pyramid. The $3 \PLH 3$ convolutions are followed by
    batch-normalization and ReLU activation. Upsampling is achieved via bilinear
    interpolation. The final predictions consist of a single $1 \PLH 1$
    convolution followed by sigmoid activation.}
  \label{fig:architecture}
\end{figure}
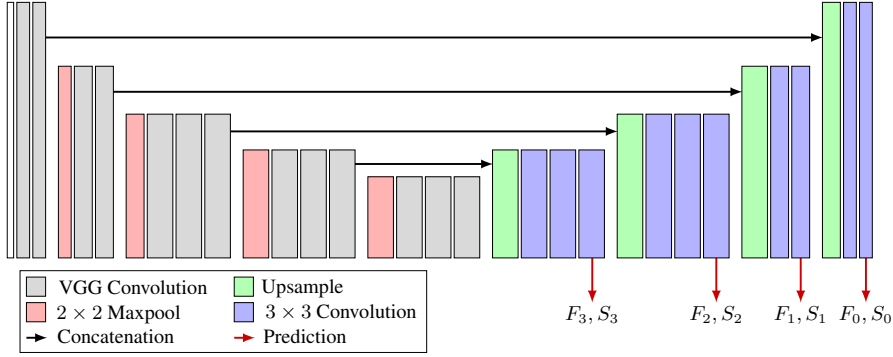

In this work, we adopt a network architecture similar to DispNet
\cite{Mayer-CVPR-2015} as it generates side predictions for constructing a
multi-level image pyramid and has proven sufficient for complex, single-view
inference tasks \cite{Zhou-CVPR-2017}. In the contractive part of the network,
we employ the learned convolutional weights of the VGG-16 network
\cite{Simonyan-ARXIV-2015}, which are not updated during training. Our second
modification involves the output predictions themselves. For a given input
image $I$, we infer a set of 16-channel feature maps $[ F^0, F^1, F^2, F^3 ]$
and single-channel saliency maps $[ S^0, S^1, S^2, S^3 ]$. Here, $F^0$ denotes
the finest resolution feature map and $F^3$ the coarsest. The same relationship
holds for saliency maps. An overview of our network architecture is shown in
Figure~\ref{fig:architecture}. How these predictions are integrated into our
relocalization framework is detailed in the next section.

\subsection{Loss Function}
\label{sec:loc-loss-function}

Our training loss comprises multiple functions over one or more $\mathbb{SE}(3)$
transforms, which are obtained through 3D registration of the inferred feature
maps. Using a simplified implementation of the inverse-composition algorithm
presented in \cite{Lv-CVPR-2018}, we ensure that this registration
process permits auto-differentiation. In contrast with \cite{Lv-CVPR-2018},
inference is performed over each image individually. Similar to their
convolutional M-estimator, our saliency maps re-weight the residuals of the
image-registration problem. However, our saliency maps are generated \emph{once}
for each image. Finally, we adopt a pure Gauss-Newton implementation, dropping
their Levenberg-Marqaurdt damping scalar.


\subsubsection{Transform Consistency}
\label{sec:loc-transform-consistency}

As previously stated, training samples consist of two references frames $r_0,
r_1$ and a query frame $q$. Each frame is represented by a rectified RGB image
and a camera projection matrix $K \in \mathbb{R}^{3\times3}$. Reference frames
are also accompanied by their respective depth maps ${D_{r_0}, D_{r_1}}$, and
relative transform $\hat{T}_{r_0,r_1} \in \mathbb{SE}(3)$.
Using the inferred feature and saliency maps, we invoke two instances of direct
3D image registration. We \textit{independently} align the query feature map
$F_q$ to each reference feature map $F_{r_0}, F_{r_1}$ to obtain the relative
transforms $T_{q,r_0}$ and $T_{q,r_1}$ by minimizing
\begin{equation}
  T_{q,r} = \argmin_{T_{q,r}} \sum_{u \in I_r}
    S_r(u) S_q(u') \Big\lVert F_r(u) - F_q(u') \Big\rVert_\gamma.
  \label{eq:photometric-error}
\end{equation}
Here $u'$ is where the coordinates $u$ in the reference image project onto the
query image, given the current estimate $T_{q,r}$, depth map $D_r$, and camera
projection matrices $K_r, K_q$. The Huber norm $\lVert \cdot \rVert_\gamma$
makes this optimization more robust to outliers. Each iteration $k$ of the
multi-level image registration yields new estimates $T^k_{q,r_0}, \,
T^k_{q,r_1}$, until we obtain the final estimates $T^*_{q,r_0}, \, T^*_{q,r_1}$.
These final estimates serve as ground-truth for the opposing alignment in the
transform consistency loss
\begin{equation}
  L_\mathrm{c} \big(T^*_{q,r_0}, \, T^k_{q,r_1}\big) =
    \Big\lVert \, \log\big( \hat{T}_{r_0, r_1} \, \big(T^k_{q, r_1}\big)^{-1} \,
    T^*_{q, r_0} \big) \, \Big\rVert_1.
  \label{consistency-loss}
\end{equation}
Intuitively, in Eq.~\eqref{consistency-loss} we are trying to compute
the difference between two estimates of the same pose. However, this first
requires moving both poses to a shared reference frame. In this case, we move
$T^k_{q, r_1}$ to reference frame $r_0$ using the provided transform
$\hat{T}_{r_0, r_1}$. We then compute the relative transform error in
$\mathfrak{se}(3)$ tangent space, and finally take the L1-norm of the resulting
residual vector.

Transform consistency eschews the need for ground-truth poses by maximizing the
agreement of two independent estimates. While there are an infinite number of
\emph{incorrect} solutions that would minimize this loss, we argue that the
solution space is largely constrained by the iterative image registration
process and a sufficiently large number of training examples. However, we
alleviate this problem further with an additional regularization term, as
described in the next section.


\subsubsection{Transform Accuracy}
\label{sec:loc-transform-accuracy}

As we assume knowledge of the relative transform $\hat{T}_{r_0, r_1}$, we can
additionally perform image registration between the two reference frames and
evaluate each intermediate transform $T^k_{r_1, r_0}$
\begin{equation}
  L_\mathrm{a}\big(T^k_{r_1,r_0}\big) = \Big\lVert \, \log\big(
    \hat{T}_{r_0, r_1} \, T^k_{r_1, r_0} \big) \, \Big\rVert_1.
  \label{accuracy-loss}
\end{equation}
The main benefit of this term is to provide stability during initialization.
Empirically, we find the network may diverge when trained with the consistency
loss alone. This is not the only benefit, however. Eq.~\eqref{accuracy-loss}
also promotes saliency maps that mask out fast moving objects and feature maps
that compensate for large camera baselines, as these challenges are still
present in a pure visual-odometry setting. Our final loss function becomes the
sum of these two terms, aggregated over each iteration of image registration,
using a constant scalar $\lambda$ to modulate the influence of each term
\begin{equation}
  \sum_k
  L_\mathrm{c}\big(T^*_{q,r_0}, T^k_{q,r_1}\big) +
  L_\mathrm{c}\big(T^k_{q,r_0}, T^*_{q,r_1}\big) +
  \lambda L_\mathrm{a}\big(T^k_{r_1,r_0}\big).
  \label{final-loss}
\end{equation}


\section{Experimental Results}
\label{sec:loc-experimental-results}

We evaluate our framework using both synthetic and real-world data. During
training and evaluation our network uses a fixed number of image-registration
iterations, with all relative transforms initialized with identity. From the
coarsest level of the image pyramid to the finest, we perform 16, 12, 8, \& 4
iterations. The input image resolution for both datasets is $640\times384$. We
train the network using the Adam Optimizer \cite{Kingma-ICLR-2014} with a
learning rate of $10^{-4}$ during initialization and $10^{-5}$ during general
training. Gradient accumulation is also used to achieve an effective mini-batch
size of 16.

To ensure the network converges reliably, we initialize the network using a
substantially higher relative transform accuracy weight $\lambda = 10$. After
training for a single epoch we lower it to $\lambda = 1$. While the transform
accuracy loss provides stability, it is the transform consistency loss that
permits us to learn robust image representations that map well across different
video sequences. To illustrate this benefit, we also evaluate our network
trained using only the transform accuracy loss, defined in
Eq.~\eqref{accuracy-loss}. We will refer to this method as ``Ours (VO)''.


\begin{figure}
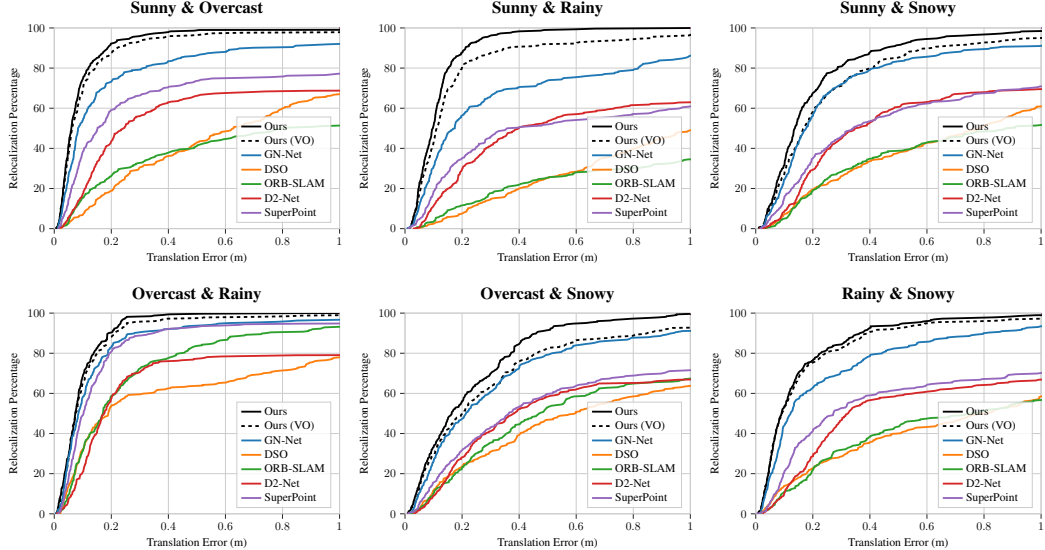

  \centering
  \begin{subfigure}{0.3275\textwidth}
    \resizebox{1.00\textwidth}{!}{\input{figures/sunnyovercast-results.tex}}
  \end{subfigure}
  \begin{subfigure}{0.3275\textwidth}
    \resizebox{1.00\textwidth}{!}{\input{figures/sunnyrainy-results.tex}}
  \end{subfigure}
  \begin{subfigure}{0.3275\textwidth}
    \resizebox{1.00\textwidth}{!}{\input{figures/sunnysnowy-results.tex}}
  \end{subfigure}\\[0.5em]
  \begin{subfigure}{0.3275\textwidth}
    \resizebox{1.00\textwidth}{!}{\input{figures/overcastrainy-results.tex}}
  \end{subfigure}
  \begin{subfigure}{0.3275\textwidth}
    \resizebox{1.00\textwidth}{!}{\input{figures/overcastsnowy-results.tex}}
  \end{subfigure}
  \begin{subfigure}{0.3275\textwidth}
    \resizebox{1.00\textwidth}{!}{\input{figures/rainysnowy-results.tex}}
  \end{subfigure}
  \caption[Cumulative relocalization accuracy plot]{Cumulative relocalization
    accuracy compared to other leading methods on the GN-Net Relocalization
    Benchmark \cite{VonStumberg-RAL-2019}. Each test consists of image pairs
    from the Robotcar dataset \cite{Maddern-IJRR-2017} that exhibit different
    weather conditions. We additionally compare the proposed system,
    \emph{Ours}, with the same framework trained only using transform accuracy
    loss, \emph{Ours (VO)}.}
  \label{fig:translation-error-plot}
\end{figure}


\subsection{Training Datasets}
\label{sec:loc-training-datasets}

For our experiments on synthetic data we employ the CARLA simulator
\cite{Dosovitskiy-CRL-17}. We simulate a car driving on rural and urban roads,
at different times of day and under different weather conditions. Each
trajectory is randomly populated with dynamic cars and pedestrians. CARLA
directly provides us with colors images, camera poses, and semantic
segmentations. However, we compute depthmaps using Semi-Global Matching
\cite{Hirschmuller-CVPR-2005} on the stereo pair, rendered with added noise. In
total, our training dataset consists of 8K images and 20K unique frame triplets
$\{r_0, r_1, q\}$, which exhibit a median camera baseline of 1.25m. We use the
same pipeline to generate a test dataset of 500 frame triplets, sampling a
completely distinct set of random trajectories.

For experiments on real-world data, we employ the Oxford Robotcar dataset
\cite{Maddern-IJRR-2017}. This dataset was captured by an autonomous car over
the course of two years, using several cameras, lidars, GPS, and IMU. Given the
extended duration of data acquisition, the captured images exhibit large changes
in visual appearance due to varying time of day, weather, and season. We select
training triplets using the global poses obtained through GPS-inertial
positioning. Again, the median camera baseline is approximately 1.25m, but can
be as high as 6m. For computing depth maps, we employ Semi-Global Matching
\cite{Hirschmuller-CVPR-2005} on the wide stereo pair. In total, our training
dataset consists of approximately 7.5K images and 15.5K unique frame triplets,
drawn from 8 Robotcar sequences.


\subsection{Experiments}
\label{sec:loc-experiments}

We first look at performance in terms of translation error. In
Figure~\ref{fig:translation-error-plot} we see how our framework performs on the
relocalization benchmark recently published with GN-Net
\cite{VonStumberg-RAL-2019}. This benchmark is separated into six different
Robotcar sequence pairs, exhibiting two distinct weather conditions (\eg sunny
and cloudy). Suitable image candidates are provided by the benchmark, allowing
frameworks to focus solely on metric pose refinement. Here we can see how our
framework compares with other leading methods including GN-Net
\cite{VonStumberg-RAL-2019}, ORB-SLAM \cite{MurArtal-TR-2017}, DSO
\cite{Engel-TPAMI-2018} SuperPoint \cite{Detone-CVPR-2018}, and D2-Net
\cite{Dusmanu-CVPR-2019}. The plots in Figure~\ref{fig:translation-error-plot}
represent cumulative relocalization accuracy. This can be interpreted as the
percentage of relocalizations (vertical axis) that are performed within a given
translation threshold (horizontal axis). A theoretically perfect system would
result in a horizontal line at the top of each plot, indicating that 100\% of
relocalizations were performed with zero error.

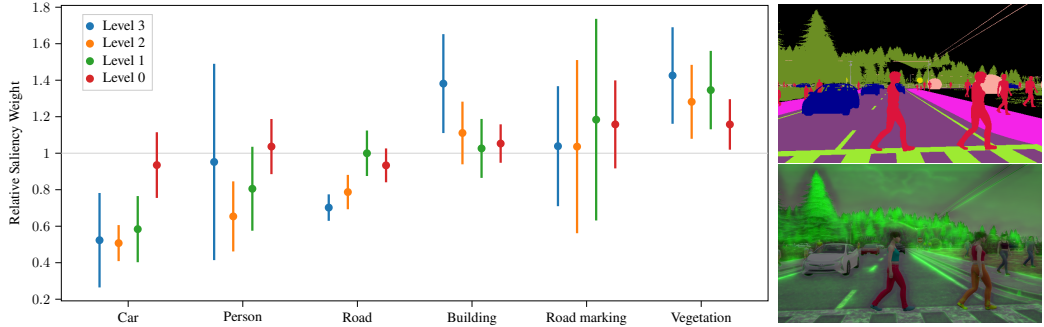
\begin{figure}
  \centering
  \begin{subfigure}{0.73\textwidth}
    \resizebox{1.00\textwidth}{!}{
\begin{tikzpicture}

\definecolor{color0}{rgb}{0.12156862745098,0.466666666666667,0.705882352941177}
\definecolor{color1}{rgb}{1,0.498039215686275,0.0549019607843137}
\definecolor{color2}{rgb}{0.172549019607843,0.627450980392157,0.172549019607843}
\definecolor{color3}{rgb}{0.83921568627451,0.152941176470588,0.156862745098039}

\begin{axis}[
legend cell align={left},
legend style={at={(0.03,0.97)}, anchor=north west, draw=white!80.0!black},
tick align=outside,
tick pos=left,
width=7.5in, height=3.5in,
x grid style={lightgray!92.02614379084967!black},
xmin=-1, xmax=36,
xtick style={color=black},
xtick={2.5,8.5,14.5,20.5,26.5,32.5},
xticklabels={Car,Person,Road,Building,Road marking,Vegetation},
y grid style={lightgray!92.02614379084967!black},
ylabel={Relative Saliency Weight},
ymin=0.190952430343259, ymax=1.80989019910794,
ytick style={color=black}
]
\addplot [semithick, white!82.74509803921568!black, forget plot]
table {%
-1 1
36 1
};
\path [draw=color0, ultra thick]
(axis cs:1,0.264540510741653)
--(axis cs:1,0.782082525956246);

\path [draw=color0, ultra thick]
(axis cs:7,0.414224026139624)
--(axis cs:7,1.48990529726283);

\path [draw=color0, ultra thick]
(axis cs:13,0.629520914949956)
--(axis cs:13,0.774603319281745);

\path [draw=color0, ultra thick]
(axis cs:19,1.11045115340172)
--(axis cs:19,1.65246306190578);

\path [draw=color0, ultra thick]
(axis cs:25,0.709762157206105)
--(axis cs:25,1.36727586387447);

\path [draw=color0, ultra thick]
(axis cs:31,1.16126906253277)
--(axis cs:31,1.69015743975396);

\path [draw=color1, ultra thick]
(axis cs:2,0.408758641585581)
--(axis cs:2,0.606121595028929);

\path [draw=color1, ultra thick]
(axis cs:8,0.461766553021862)
--(axis cs:8,0.84573900479422);

\path [draw=color1, ultra thick]
(axis cs:14,0.693003148855545)
--(axis cs:14,0.881243428021462);

\path [draw=color1, ultra thick]
(axis cs:20,0.939173072288813)
--(axis cs:20,1.28256927262037);

\path [draw=color1, ultra thick]
(axis cs:26,0.561632574809223)
--(axis cs:26,1.51053010554518);

\path [draw=color1, ultra thick]
(axis cs:32,1.07905272556255)
--(axis cs:32,1.48417703779156);

\path [draw=color2, ultra thick]
(axis cs:3,0.402728788497349)
--(axis cs:3,0.765174694566371);

\path [draw=color2, ultra thick]
(axis cs:9,0.575299513915854)
--(axis cs:9,1.03527437520549);

\path [draw=color2, ultra thick]
(axis cs:15,0.875098167632184)
--(axis cs:15,1.12430148895827);

\path [draw=color2, ultra thick]
(axis cs:21,0.86481130368593)
--(axis cs:21,1.18779725304385);

\path [draw=color2, ultra thick]
(axis cs:27,0.631509810954818)
--(axis cs:27,1.73630211870954);

\path [draw=color2, ultra thick]
(axis cs:33,1.13111368364019)
--(axis cs:33,1.56071671971955);

\path [draw=color3, ultra thick]
(axis cs:4,0.755053488537468)
--(axis cs:4,1.11489760772721);

\path [draw=color3, ultra thick]
(axis cs:10,0.88531372784232)
--(axis cs:10,1.18753437909796);

\path [draw=color3, ultra thick]
(axis cs:16,0.840675018595194)
--(axis cs:16,1.02620168548067);

\path [draw=color3, ultra thick]
(axis cs:22,0.947662238500455)
--(axis cs:22,1.15823889243455);

\path [draw=color3, ultra thick]
(axis cs:28,0.916844680733839)
--(axis cs:28,1.39902490644011);

\path [draw=color3, ultra thick]
(axis cs:34,1.01940711842848)
--(axis cs:34,1.29593645828472);

\addplot [ultra thick, color0, mark=*, mark size=2, mark options={solid}, only marks]
table {%
1 0.52331151834895
7 0.952064661701228
13 0.70206211711585
19 1.38145710765375
25 1.03851901054029
31 1.42571325114336
};
\addlegendentry{~~Level 3}
\addplot [ultra thick, color1, mark=*, mark size=2, mark options={solid}, only marks]
table {%
2 0.507440118307255
8 0.653752778908041
14 0.787123288438504
20 1.11087117245459
26 1.0360813401772
32 1.28161488167706
};
\addlegendentry{~~Level 2}
\addplot [ultra thick, color2, mark=*, mark size=2, mark options={solid}, only marks]
table {%
3 0.58395174153186
9 0.805286944560673
15 0.999699828295227
21 1.02630427836489
27 1.18390596483218
33 1.34591520167987
};
\addlegendentry{~~Level 1}
\addplot [ultra thick, color3, mark=*, mark size=2, mark options={solid}, only marks]
table {%
4 0.93497554813234
10 1.03642405347014
16 0.93343835203793
22 1.0529505654675
28 1.15793479358698
34 1.1576717883566
};
\addlegendentry{~~Level 0}
\end{axis}

\end{tikzpicture}}
  \end{subfigure}
  \begin{subfigure}{0.25\textwidth}
    \includegraphics[width=\textwidth,height=0.60\textwidth,
    trim=20 0 20 0, clip]{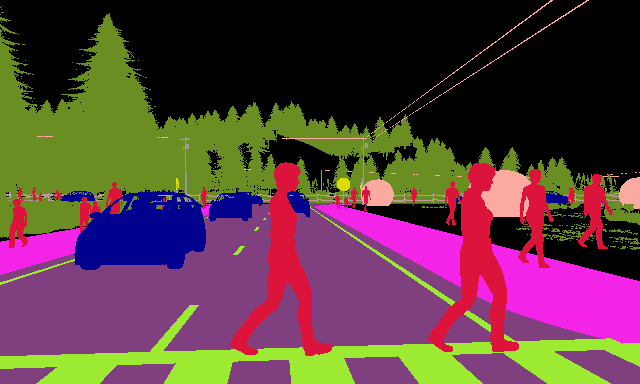}\\
    \includegraphics[width=\textwidth,height=0.60\textwidth,
    trim=20 0 20 0, clip]{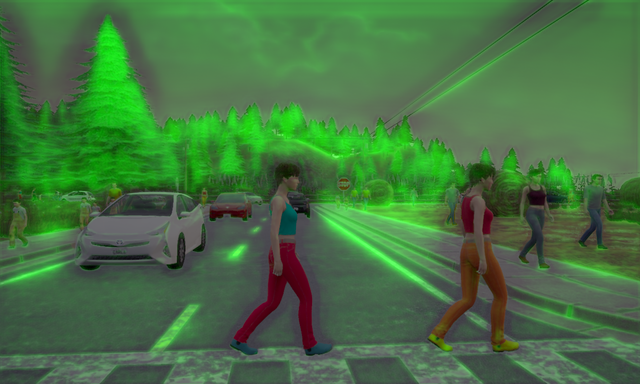}
  \end{subfigure}
  \caption[Relative saliency weights]{Using the CARLA simulator
    \cite{Dosovitskiy-CRL-17}, we generate a dataset of images with their
    respective semantic labels (\emph{top right}). We then evaluate our saliency
    maps (\emph{bottom right}) for each class in the Cityscape taxonomy,
    comparing the resulting weights with those obtained using a uniform
    weighting strategy (\emph{left}). The mean and standard-deviation of each
    pyramid level is plotted. Values above the gray line suggest additional
    focus while those below suggest the class is being ignored.}
  \label{fig:relative-saliency-plot}
\end{figure}

Next, we employ the semantic segmentations obtained using the CARLA simulator
\cite{Dosovitskiy-CRL-17} to directly evaluate our saliency maps. Of particular
interest is how well our saliency maps mask out dynamic objects (\eg cars and
pedestrians). For each image in our test dataset, we compare the inferred
saliency weights with those obtained using a uniform weighting strategy. As an
example, if a instance segmentation occupies 50\% of the image, but only
accounts for 25\% of the total saliency weight, its \emph{relative saliency
weight} is 0.5. Values above 1.0 suggest additional focus is being directed
towards the class, while values below 1.0 suggest the class is being ignored. We
perform this analysis independently on each pyramid level using the Cityscape
taxonomy \cite{Cordts-CVPR-2016}. The observed relative saliency weight means
and standard-deviations are plotted in Figure~\ref{fig:relative-saliency-plot}.


Image pairs in the GN-Net Relocalization Benchmark primarily exhibit changes in
weather with moderate initial camera baselines. We evaluate our framework
further with a more challenging dataset comprising day-night and seasonal
changes, with significantly larger camera baselines. Aligning lidar pointclouds
to obtain ground-truth, inter-sequence transforms, as performed in
\cite{VonStumberg-RAL-2019}, is not only labor intensive but also sensitive to
the hyperparameters and initial pose. We therefore adopt a metric similar to
relative pose error \cite{Konolige-RR-2010, Geiger-CVPR-2012}
\begin{equation}
  E =
    \big\lVert
      \mathrm{translation}\big(
        \hat{T}_{r_0, r_1}(T^*_{q, r_1})^{-1}T^*_{q, r_0}
      \big)
    \big\rVert.
  \label{eq:relative-pose-error}
\end{equation}
Intuitively, we are comparing the relative translation computed via
intra-sequence visual-odometry, with that computed from two independent map
registrations. Our test dataset contains 40K frame triplets, with 100 samples
drawn from 400 Robotcar sequence-pairs. Sampling in this fashion allows us to
build a ``confusion matrix'' indicating how well reference frames from one
sequence align the query frames of another. For this experiment, we retrain our
network with a dataset of similar size and scope, created from a distinct set of
Robotcar sequences. Results are shown in Figure \ref{fig:relative-pose-error}.


\section{Discussion}
\label{sec:loc-discussion}

As shown in Figure~\ref{fig:translation-error-plot}, our approach consistently
outperforms other leading methods. Compared to the next top performer, we
successfully relocalize within 25cm for 15.31\% more of the benchmark. We
attribute the observed performance gains to our network architecture and image
registration framework, as even the \emph{Ours (VO)} approach proves to be quite
accurate. By training directly on the 3D registration problem and employing
multi-level saliency maps, we not only learn to ignore dynamic objects but also
to down-weight objects in the foreground. In general these will exhibit larger
optical flow vectors, which are problematic in direct image registration. By
giving these objects less significance in the coarser saliency maps we can
overcome large initial camera baselines.


\begin{figure}
  \centering
  \begin{subfigure}{\textwidth}
    \resizebox{\textwidth}{!}{\input{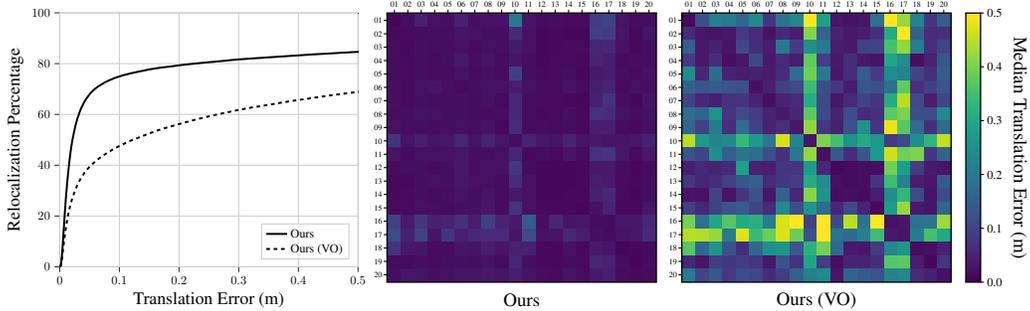}}
  \end{subfigure}
  \caption[Relative pose error]{Cumulative relocalization accuracy (\emph{left})
    and ``confusion matrices'' (\emph{right}) comparing the relative translation
    errors of the propose method, \emph{Ours}, with the same framework trained
    only using transform accuracy loss, \emph{Ours (VO)}. Each column of a
    matrix represents the Robotcar sequence of the reference frames, while each
    row represents the sequence of the query frames. The color of a cell
    indicates the median translation error, computed over 100 triplet samples.}
  \label{fig:relative-pose-error}
\end{figure}

In this first experiment, our network is trained with eight sequences of the
Robotcar dataset. The GN-Net Relocalization Benchmark
\cite{VonStumberg-RAL-2019}, however, publishes only two such training
sequences. We argue this remains a valid comparison, as our solution does not
leverage \emph{any} ground-truth, inter-sequence transforms. In fact, the main
benefit of this approach is the ease in which training data can be acquired.
Given the minimal requirements outlined in Section \ref{sec:method}, we can
incorporate additional sequences only knowing the car's global position within
6m and heading within 15 degrees.

Figure~\ref{fig:relative-saliency-plot} shows that our network learns to ignore
dynamic objects, consistently down-weighting the classes ``car'' and ``person''.
In contrast, we see that larger, static objects receive additional attention,
which aligns well with our intuition. As CARLA does not simulate seasonal
changes in foliage, we also see the network safely learns to rely on vegetation.
We do note, however, that higher resolution saliency maps trend towards a
uniform weighting strategy (indicated by the gray line in
Figure~\ref{fig:relative-saliency-plot}). We suspect that this is due to the
convergence basin becoming narrower at each subsequent level, such that dynamic
objects have significantly less impact on the refinement of the final pose, and
are likely handled by the Huber norm used in Eq.~\eqref{eq:photometric-error}.

In Figure~\ref{fig:translation-error-plot}, training our network with transform
consistency loss appears to result in moderate performance gains, as compared to
training with transform accuracy loss alone. However, the true benefit of using
transform consistency is revealed when evaluating on a more challenging
relocalization dataset, as seen in Figure~\ref{fig:relative-pose-error}. Most
notably, we see that the \emph{Ours (VO)} method fails to register nighttime
sequences (10, 16, and 17) with daytime sequences. In fact, the majority of
off-diagonal entries in its confusion matrix indicate significantly higher
translation error. This can be expected, as the network does not learn from
inter-sequence training examples. Evaluating with the relative pose error
\eqref{eq:relative-pose-error} also shows our framework is more accurate than
the plots in Figure~\ref{fig:translation-error-plot} would suggest.

In the most extreme conditions, we suspect that our static saliency maps,
inferred once from a single input image, will not perform as well as those
iteratively updated during image registration, as proposed in
\cite{Lv-CVPR-2018}. However, the benefit of our approach is that the saliency
maps are not dependent on the current image pair. Consequently, they may be
evaluated in isolation to determine which frames are suitable map keyframes.
For example, we could reject frames dominated by dynamic objects and other
features the network has learned are unreliable, as indicated by low-activation
in the saliency maps. This is a potential direction for future research.


\section{Conclusion}
\label{sec:loc-conclusion}

We have presented a novel training loss, which permits unsupervised learning of
dense feature and saliency maps for robust metric relocalization. This greatly
reduces the constraints on dataset acquisition, allowing data to be be captured
\textit{in situ} by mobile agents with little additional overhead. Our framework
attains state-of-the-art results on the GN-Net Relocalization Tracking
Benchmark, significantly outperforming other leading methods. These performance
gains are attributed to our multi-resolution saliency maps and the larger
training datasets that unsupervised learning affords.


\bibliography{corl2020}
\end{document}